\documentclass{article}

\usepackage[preprint,nonatbib]{neurips_2022}

\usepackage[utf8]{inputenc} 
\usepackage[T1]{fontenc}    
\usepackage{url}            
\usepackage{booktabs}       
\usepackage{amsfonts}       
\usepackage{nicefrac}       
\usepackage{microtype}      
\usepackage{xcolor}         

\usepackage{amsmath}
\usepackage{amssymb}
\usepackage{wrapfig}
\usepackage{xcolor}

\usepackage{microtype}
\usepackage{graphicx}
\usepackage{subfigure}
\usepackage{multirow}
\usepackage{makecell}
\usepackage{booktabs} 

\definecolor{citecolor}{HTML}{248B21}
\usepackage[pagebackref,breaklinks,citecolor=citecolor,colorlinks]{hyperref}

\newcommand{\aps}{AP & AP$_{50}$ & AP$_{75}$}

\newcommand{\etal}{\textit{et al}. }
\newcommand{\ie}{\textit{i}.\textit{e}.}
\newcommand{\eg}{\textit{e}.\textit{g}.}
\newcommand{\wrt}{\textit{w}.\textit{r}.\textit{t}.}
\newcommand{\tline}{\Xhline{1pt}}

\newcommand{\model}{\mbox{\sc{Featurized QR-CNN}}}
\newcommand{\cascademodel}{\mbox{Cascade \sc{Featurized QR-CNN}}}

\title{Featurized Query R-CNN}

\author{
Wenqiang Zhang $ ^{1}{}^{*}$
~~~~~~~
Tianheng Cheng $ ^{1} $\thanks{Equal contribution (\texttt{\{wq\_zhang, thch\}@hust.edu.cn}). This work was done when Wenqiang Zhang, Tianheng Cheng, and Shaoyu Chen were interning at Horizon Robitics.}
~~~~~~~
Xinggang Wang $ ^{1} $\thanks{Corresponding author (\texttt{xgwang@hust.edu.cn}).}
~~~~~~~
Shaoyu Chen $ ^{1} $\\[0.152cm]
\textbf{Qian Zhang} $ ^{2} $
~~~~~~~
\textbf{Wenyu Liu} $ ^{1} $ \\[0.152cm]
$^1$ School of EIC, Huazhong University of Science \& Technology ~~~
$^2$ Horizon Robotics }

\begin{document}

\maketitle

\begin{abstract}
  The query mechanism introduced in the DETR method is changing the paradigm of object detection and recently there are many query-based methods have obtained strong object detection performance. However, the current query-based detection pipelines suffer from the following two issues. Firstly, multi-stage decoders are required to optimize the randomly initialized object queries, incurring a large computation burden. Secondly, the queries are fixed after training, leading to unsatisfying generalization capability. To remedy the above issues, we present featurized object queries predicted by a query generation network in the well-established Faster R-CNN framework and develop a Featurized Query R-CNN. Extensive experiments on the COCO dataset show that our Featurized Query R-CNN obtains the best speed-accuracy trade-off among all R-CNN detectors, including the recent state-of-the-art Sparse R-CNN detector. The code is available at \url{https://github.com/hustvl/Featurized-QueryRCNN}.
\end{abstract}

\section{Introduction}

Object detection as the fundamental task in computer vision aims to localize and recognize objects in images, which drives massive applications, \eg, autonomous vehicles and robotics. The deep convolutional neural networks have rapidly boosted the developments of object detection.
It is remarkable that the region-based methods~\cite{girshick2015fast,ren2015faster}, \eg, Faster R-CNN~\cite{ren2015faster}, have made great progress and achieved excellent results on large-scale benchmarks, \eg, MS-COCO~\cite{COCO}.

Recently, Sun \etal reformulate the region-based methods by replacing the region proposal network (RPN~\cite{ren2015faster}) with a fixed set of learnable object proposals (boxes and queries) and propose an end-to-end framework, \ie, Sparse R-CNN~\cite{sun2020sparse}.
It adopts sparse object proposals to extract region features through RoI-Align~\cite{he2017mask} and updates these object proposals through 6 iterative region-based decoders.
Specifically, the decoder in Sparse R-CNN comprises the self-attention to interact with other object proposals and a dynamic head to update proposals from region features.
In addition, Sparse R-CNN adopts bipartite matching for label assignment and serves as a purely end-to-end approach without non-maximum suppression (NMS).
Empirically, the learnable proposals are image-agnostic and represent potential object locations summarized through training across the dataset, which require multiple iterative decoders for image-aware refinement and attend to objects in individual images.
Notably, reducing the iterative stages severely degrades the performance since the proposals are under-optimized for localization and recognition.
Compared to classical R-CNN detectors \cite{ren2015faster,CascdeCaiV21}, Sparse R-CNN has much heavier decoders and more stages, which incur larger computation complexity and higher inference latency, limiting the practical applications.

In this paper, we present the \textbf{featurized queries} which are initialized through the image content to alleviate the above issue. 
We empirically observe that using object features from images as object queries instead of learnable queries can considerably reduce the stages of decoders without performance drop.
Specifically, we propose a query generation network (QGN) to generate image-aware object queries along with object boxes, which is motivated by RPN~\cite{ren2015faster}. 
Query generation network designed in an anchor-free style aims to localize foreground objects by bounding boxes and provide object queries.
Compared to RPN, the proposed QGN does not require complicated post-processing, \eg, NMS and  proposal sampling, which is rather efficient for query generation.

\begin{figure*}
\centering
\vspace{-2pt}
\includegraphics[width={1\textwidth}]{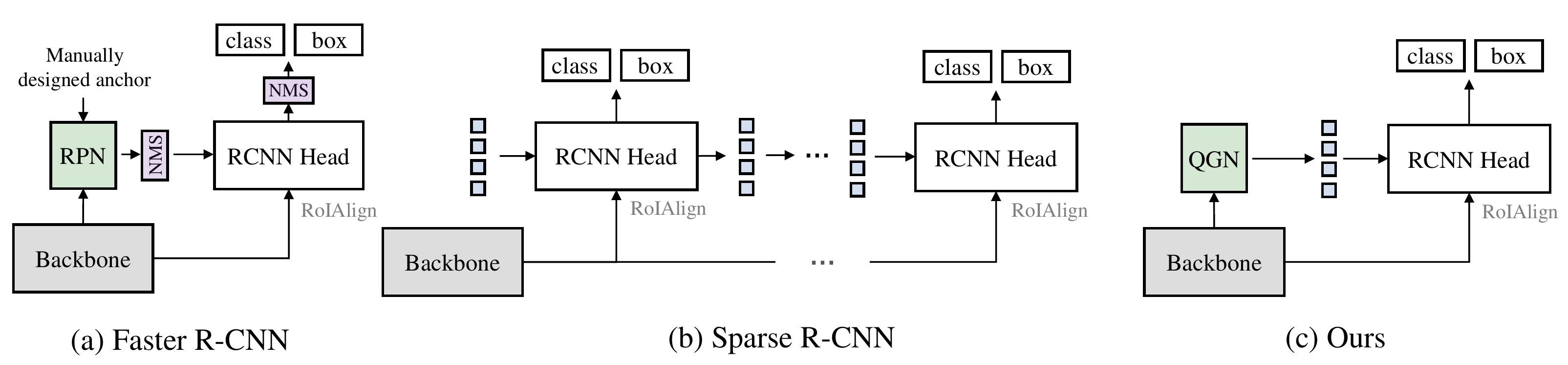}
\vspace{-10pt}
\caption{\textbf{Structure comparison.} Compared with Faster R-CNN, we remove the NMS during RPN and R-CNN stage, achieving a faster inference speed. Compared with Sparse R-CNN, we propose QGN to alleviate the dependence of iterative structures.}
\label{fig: stru_cmp}
\end{figure*}

\begin{wrapfigure}[18]{r}{0.48\linewidth}
\vspace{-10pt}
\centering
\includegraphics[width=\linewidth]{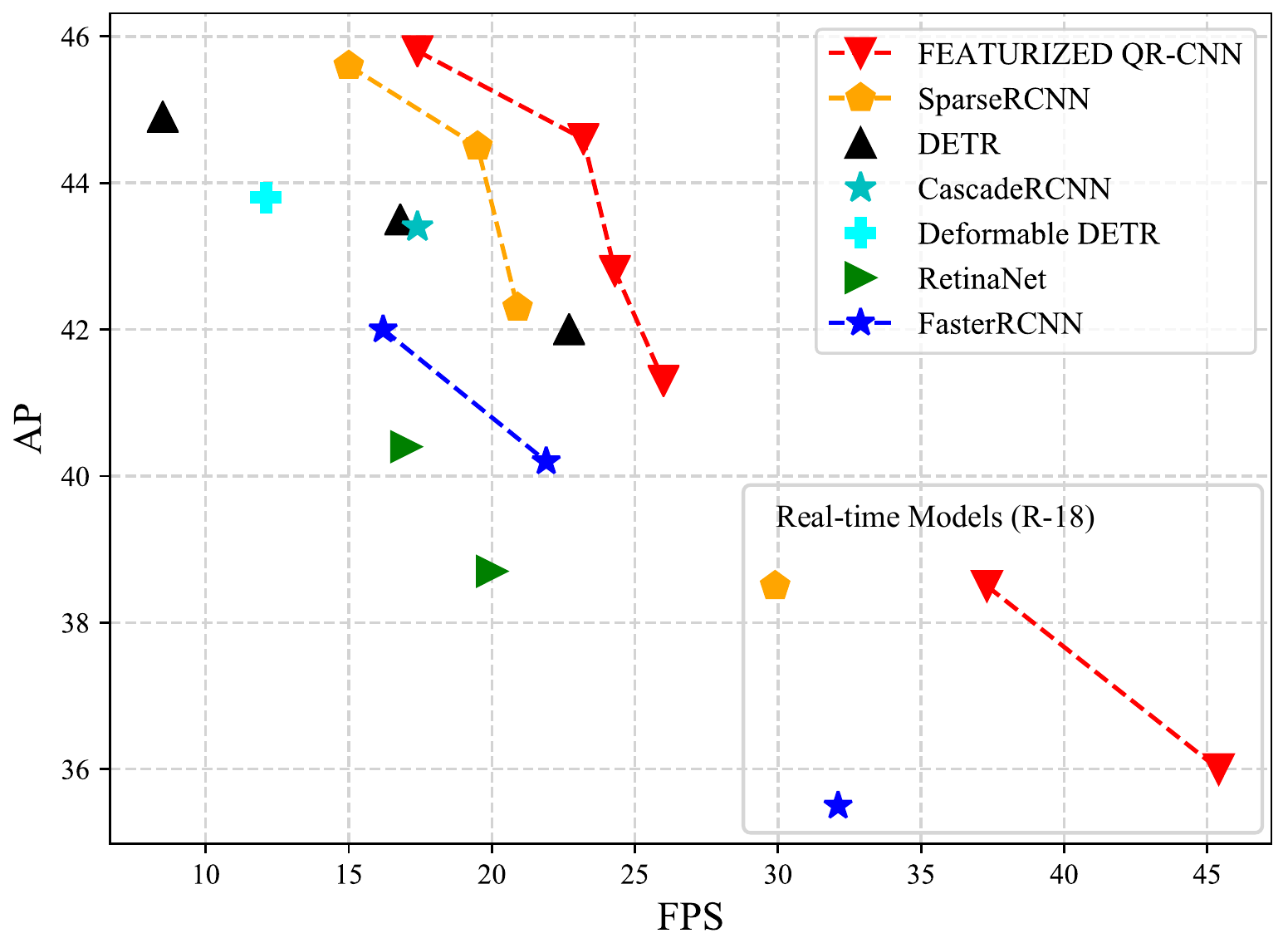}
\vspace{-20pt}
\caption{\textbf{Speed and Accuracy Trade-off.} The proposed \model{} achieves better performance and higher inference speed than the state-of-the-art methods. Inference speeds are measured using one NVIDIA 2080Ti GPU.}
\label{fig: inf_time}
\end{wrapfigure}
Further, we leverage the QGN to provide object queries and adopt one region-based decoder to refine the object proposals. Fig.~\ref{fig: stru_cmp} shows the structure comparison with RPN-based detectors (\ie, Faster R-CNN), query-based detectors (\ie, Sparse R-CNN) and our detector. QGN builds the connection between previous RPN-based dense object detectors and the query-based sparse object detectors. The QGN-based featurized query R-CNN, termed \model, takes the advantages of both dense and sparse detectors, \ie, it performs purely end-to-end detection and does not need NMS, and it only requires very a few self-attention-based decoders layers and computationally fast. Thus, our \model~is both accurate and fast. 

To demonstrate the effectiveness of the featurized query design, we conduct massive experiments on the MS-COCO~\cite{COCO} benchmark. Without bells and whistles, \model{} achieves superior speed-accuracy trade-off than the well-established Faster R-CNN and Sparse R-CNN as well as other counterparts, as shown in Fig.~\ref{fig: inf_time}.
Besides, the extensive experiments also demonstrate the fast convergence and generalization capability of the proposed featurized queries along with the proposed query generation network.
\model{} retains the two-stage paradigm of object detectors and improves the inference speed and accuracy simultaneously, which can serve as a new baseline for object detection.

\section{Related Work}

\subsection{Region-based Object Detectors}
The development of deep convolutional networks have rapidly boosted the performance of modern object detection. 
\cite{PichGirshick14,girshick2015fast} present a region-based paradigm for object detection and adopt convolutional networks to extract region-of-interest (RoI) features with pre-computed object proposal boxes~\cite{selective_search}.
Ren et.al propose region proposal networks (RPN) ~\cite{ren2015faster} to get rid of pre-computed proposals in Fast R-CNN~\cite{girshick2015fast} and present Faster R-CNN~\cite{ren2015faster}, which serves as a well-known strong baseline among region-based detectors.
He et.al. propose Mask R-CNN~\cite{he2017mask}, which incorporates a segmentation head into Faster R-CNN for instance segmentation.
Several methods~\cite{CascdeCaiV21,chen2019hybrid,guided_anchor,cascade_rpnuJPY19} based on Faster R-CNN leverage cascade architectures to improve the quality of the bounding boxes.

\subsection{Query-based Object Detectors}

Recently, query-based methods~\cite{DETRCarionMSUKZ20} bring a new paradigm to object detection. 
DETR~\cite{DETRCarionMSUKZ20}, as the first query-based detector, integrates the transformer encoder-decoder architecture upon the convolutional networks and adopts a set of object queries to decode object bounding boxes.
Deformable DETR~\cite{DeformableZhu20} proposes multi-scale deformable attention, motivated by \cite{dai2017deformable}, and achieves better performance and faster convergence speed.
YOLOS~\cite{fang2021you} proposes a pure sequence-to-sequence query-based detection scheme and shows strong detection performance can be obtained without using the convolutional backbone.
Several works~\cite{gao2021fast,conditionaldetrMengCFZLYS021,wang2021anchor,dabdetr} also alleviate the slow convergence problem of DETR through incorporating position information. 
\cite{EfficientYao21,roh2021sparse} based on Deformable DETR adopt auxiliary supervisions to prune the image features to lower the computation cost in transformer encoders or decoders, and achieve good trade-off between accuracy and computation budget.
Sparse R-CNN~\cite{sun2020sparse} defines a sparse set of learnable queries and boxes, and adopts 6 region-based decoders to refine bounding boxes.
Sparse R-CNN simplifies the two-stage detectors and achieves excellent results. QueryInst \cite{fang2021instances} shows Sparse R-CNN can be extended for high-quality instance segmentation.
However, the decoders in Sparse R-CNN are much heavier than the vanilla fully-connected head in Faster R-CNN, which incurs large computation cost and limits the inference speed.

\section{Method}
\subsection{Revisiting Query-based Methods}
Recently, query-based methods \cite{DETRCarionMSUKZ20,sun2020sparse} reformulate object detection as a set prediction problem and adopt a fixed-size set of object queries to detect objects.
The set prediction loss enforces the one-to-one assignment between the object queries and the ground-truth objects, which enables the end-to-end object detection without non-maximum suppression (NMS).
In addition, query-based methods require multiple iterative decoders with self-attention to refine the object queries, \eg, 6 transformer decoders in DETR.

\paragraph{Set Prediction Loss} 
The set prediction loss aims to finding a one-to-one assignment between the object queries and ground-truth objects through Hungarian algorithm~\cite{stewart2016end}, and then calculate the loss according to the assignment.
Let $y=\{y_i\}_{y=1}^{M}$ denote the ground truth of a set of objects, and $\hat{y}=\{\hat{y_i}\}_{i=1}^{N}$ denote a set of predictions. In general, $N{\gg}M$, we pad $y$ with $\varnothing$ (no object) to match the size of $\hat{y}$. The target is to find a bipartite matching ${\sigma \in \Phi}$ between these two sets.
\begin{equation}
    \hat{\sigma} = \mathop{\arg\min}_{\sigma \in \Phi}\sum_{i=1}^{N}{L_{match}(y_{i}, \hat{y}_{\sigma(i)})},
\end{equation}
where $L_{match}$ is the matching cost between ground truth $y_{i}$ and prediction $\hat{y}_{\sigma(i)}$ with index $\sigma(i)$.  

\paragraph{Iterative Decoders for Query Refinement}
Query-based methods define a sparse set of image-agnostic learnable object queries~\cite{DETRCarionMSUKZ20},
which imply the potential locations of the objects.
To obtain the image-aware detection results, query-based methods adopt iterative decoders to progressively update the object queries.    
Specifically, each decoder comprises a self-attention module to exchange information among object queries and a cross-attention~\cite{DETRCarionMSUKZ20} or dynamic convolution~\cite{sun2020sparse} to aggregate contextual information from image features. 
Notably, the iterative decoders are critical for query-based methods to refine the image-agnostic queries to recognize and localize the objects in the input image.
Reducing the number of decoders will degrade the performance rapidly.
However, the heavy decoders will bring large computation burden, especially for smaller backbones, \eg, the decoders of Sparse R-CNN take 58.4\% of the inference time with ResNet-18 as the backbone.

\subsection{Featurized Query R-CNN}

\begin{figure}[]
    \centering
    \includegraphics[width=0.90\linewidth]{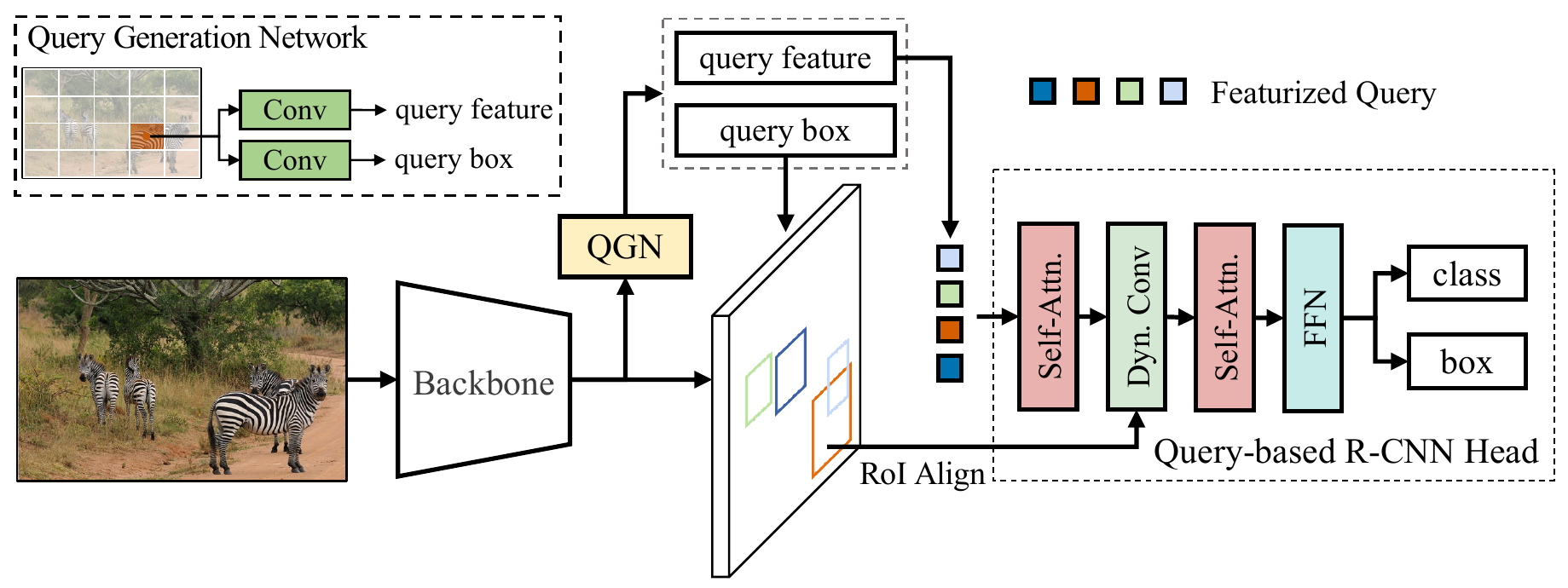}
    \caption{\textbf{The overall framework of FEATURIZED QR-CNN}. \model{} adopts a query generation network (QGN) to generate a sparse set of query boxes along with query features based on the image features. Then the query features and the RoI features (from query boxes) will be fed into the R-CNN head for classification and box regression.}
    \label{fig: overall_fig}
\end{figure}

In this paper, we propose image-aware object queries, \ie, featurized queries, to reduce the refinement iterations to simplify the query-based method for faster inference speed.
Fig.~\ref{fig: overall_fig} illustrates the overall framework of the proposed \model, which inherits the general framework of main-stream R-CNN based methods, \eg, Faster R-CNN. 
The simple QGN (Query Generation Network) motivated by RPN~\cite{ren2015faster} generates a sparse set of proposals and featurized queries in a one-to-one correspondence manner. 
Then the newly-designed R-CNN head~\cite{sun2020sparse} extracts the RoI features and performs a cross-attention to extract object features. 
The final predictions are directly obtained by a feed forward network (FFN) same as ~\cite{DETRCarionMSUKZ20, sun2020sparse}.

\subsubsection{Query Generation Network}

\paragraph{Architecture} The query generation network based on the anchor-free detector, is built upon the top of FPN~\cite{DBLP:conf/cvpr/LinDGHHB17}.
QGN adopts a dense head with a $3\times3$ convolutional layer followed by several $1\times1$ convolutions for objectness prediction, query box regression, and query feature.
Let $F_l \in \mathbb{R}^{H_l\times{W_l}\times{C_l}}$ be the feature maps at level $l$ of FPN. For each location $(x, y)$ on the feature map $F_{l}$, the network predicts an objectness score and box offsets.
The objectness score is category-agnostic and aims selecting the set of object queries (query features and query boxes). 
As for bounding box regression, we directly regress the distances from the pixel center to the four sides of the bounding boxes, \ie, ($l$, $t$, $r$, $b$).
By defaults, we use the features from P3 to P7\footnote{P$x$ refers to the FPN features with $\frac{1}{2^x}$ size of input images.}.
Besides, we additionally add a branch (a simple $1\times1$ convolution) to generate object queries $Q_l\in \mathbb{R}^{H_l\times{W_l}\times{D_l}}$, where the $D_l$ is the dimension. 
We provide more analysis on the design of query generation network in experiments.

\paragraph{Featurized Query Generation} 
In order to generate a fixed-size set of featurized object queries along with the query boxes, we first gather the pixel-wise predictions across all levels as the object candidates.
Then we simply apply the top-K (K is set to 100) selection according to the objectness scores over all predictions.
Correspondingly, we gather the query features for each selected prediction from image features $\{Q_l\}$. 



\paragraph{Training Query Generation Network} 
\label{qgn_loss}
To enforce the query generation network to output the sparse set of object queries, we also adopt the bipartite matching to assign ground truths to the predictions.
We first combine the the objectness scores and quality of the predicted query boxes, and define the matching function inspired by \cite{EndWang20} as follows as:
\begin{equation}
    {\mathcal{Q}_{match}} = {\mathcal{Q}_{obj.}}^{1-\alpha} \cdot {\mathcal{Q}_{IoU}}^{\alpha},
\end{equation}
where $\alpha$ is coefficient and $\mathcal{Q}_{IoU}$ is the IoU quality. Inspired by the implementation of ~\cite{tian2019fcos}, the quality function only considers the points located in the GT boxes. As for the loss function, we formulate it as follows:
\begin{equation}
    {\mathcal{L}_{loss}} = {\lambda_{obj.}} \cdot \mathcal{L}_{obj.} + {\lambda_{giou}} \cdot {\mathcal{L}_{giou}},
\end{equation}
where the $\mathcal{L}_{obj.}$ is focal loss and $\mathcal{L}_{giou}$ is the GIoU loss, $\lambda_{giou}$,  $\lambda_{obj.}$ are coefficients. 

\subsubsection{Query-based R-CNN Head}

The design of query-based R-CNN Head generally conforms to the standard structure of transformer-decoder~\cite{AttentionVaswani17}.
As shown in Fig.~\ref{fig: overall_fig}, we firstly extract the region features according to the query boxes through RoI-Align~\cite{he2017mask}.
And the featurized queries are fed into the multi-head self-attention module to interaction among objects.
Then a dynamic convolution with weights generated by featurized queries will be applied upon region features for interaction between object queries and region-wise image features.
In addition, we add an extra RoI-level self-attention module to further enhance the interaction among object queries, which matters much when we reduce the number of decoders in \cite{sun2020sparse}.

\subsubsection{Cascade Refinement}
\begin{figure}[thbp]
\centering
\includegraphics[width=0.6\linewidth]{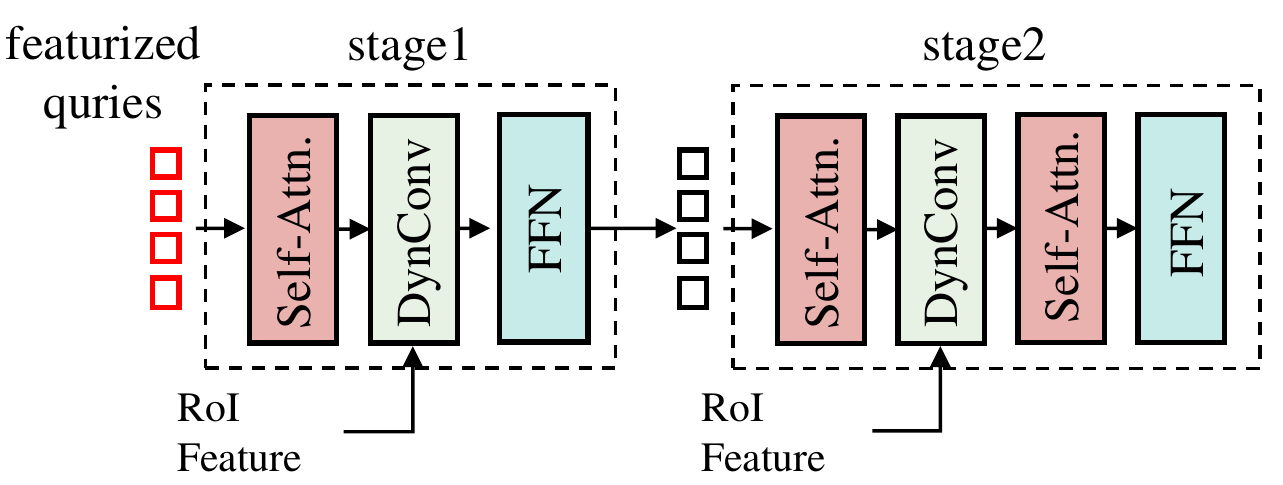}
\caption{\textbf{Cascade Head.} We insert a standard decoder~\cite{sun2020sparse} containing a self-attention and dynamic convolution to further refine the queries.}
\label{fig: double_head}
\end{figure}

To achieve better performance, we extend the \model{} with a cascade head, as shown in Fig.~\ref{fig: double_head}.
The cascade head aims to further refining the box predictions and consists of a query-based R-CNN head and a standard decoder~\cite{sun2020sparse}.
Different from previous methods using 6 or more decoders~\cite{DETRCarionMSUKZ20,DeformableZhu20,sun2020sparse} to make the image-agnostic object queries attend to objects, the proposed \model{} can generate image-aware query boxes and requires fewer decoders to refine the bounding boxes.
In \model{}, using a cascade head with an extra decoder can bring better trade-off between speed and accuracy, which is further discussed in Sec.~\ref{sec:num_deocders}.

\section{Experiments} 

In this section, we conduct experiments on the challenging COCO~\cite{COCO} dataset, which includes $80$ categories and contains 118k, 5k, and 20k images for training, validation, and testing.
We train all models on \texttt{train2017} and evaluate the models on \texttt{val2017} unless specified.

\subsection{Experimental Settings}
Our implementation is based on Detectron2~\cite{wu2019detectron2} and PyTorch. Following \cite{sun2020sparse}, the multi-scale training and $270k$ scheduler are used, the initial learning rate is set to $2.5\times10^{-5}$, divided by $10$ at $210$k iterations and $250k$ iterations. We use AdamW~\cite{loshchilov2017decoupled} optimizer with weight-decay $1\times10^{-4}$. The backbone is initialized with the pretrained weights on ImageNet~\cite{imagenet}. For fair comparison, the label assignment procedures and data augmentation operations follow the default setting of~\cite{sun2020sparse}. By defaults, we use \model~ with $100$ queries and ResNet-50-FPN~\cite{he2016deep} for ablation experiments.

During testing, images are resized to maximum scale of $1333\times800$. The inference speed reported is measured using a single NVIDIA 2080ti GPU. For the sparse proposal generation part, we directly get the top $K$ output prediction according to the corresponding objectness scores. For R-CNN head part, we get top $100$ prediction boxes with their scores without post processing. The simple processing framework makes our inference speed very fast.

\begin{figure*}[t]
\centering
\includegraphics[width=\linewidth]{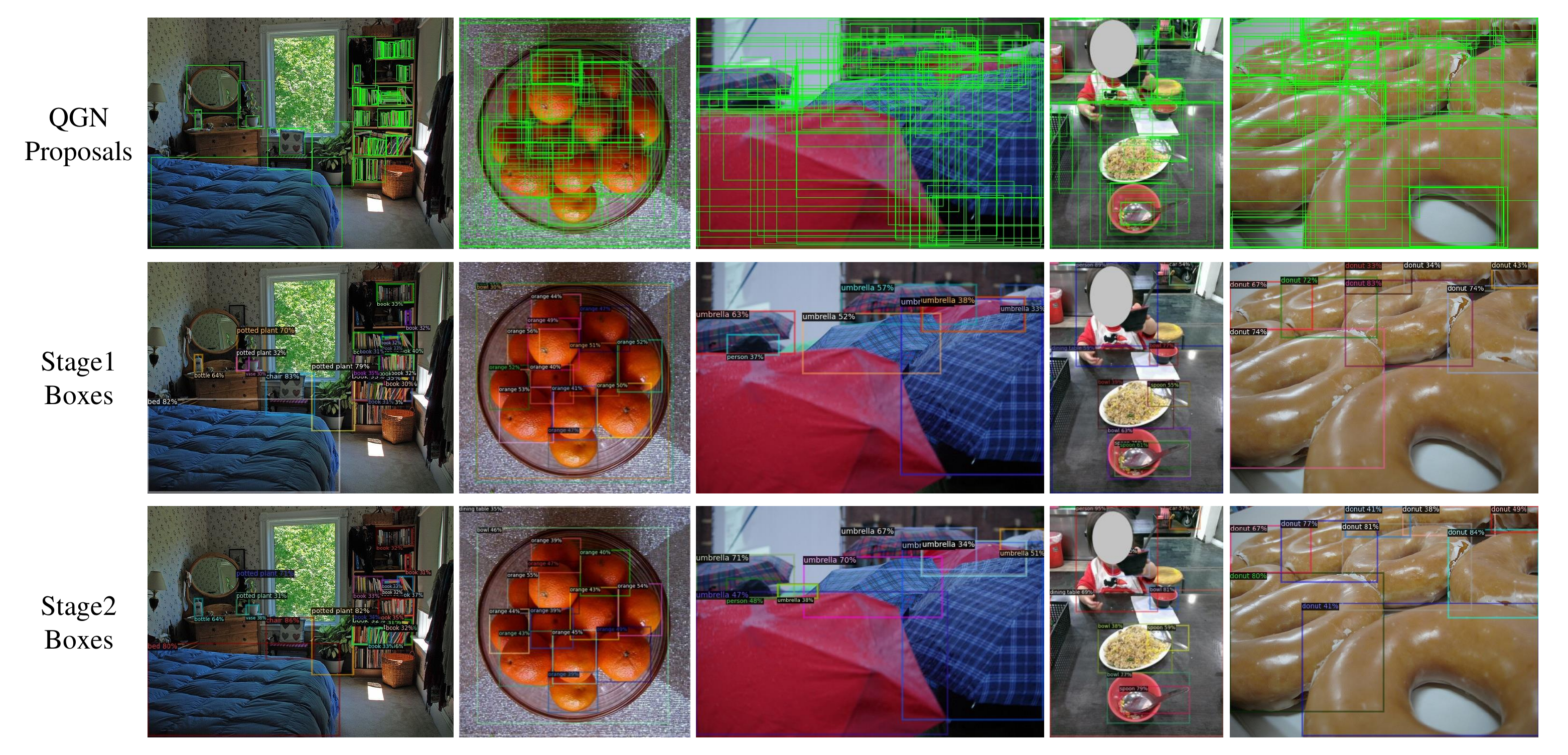}
\vspace{-0.8cm}
\caption{Visualizations of the bounding boxes of different stages. The proposed QGN yields the proposal boxes which are well-localized on foreground objects. The cascade two stages reduce the redundant predictions and refine the boxes for accurate localization.}
\label{fig: rpn_proposals}
\end{figure*}

\subsection{Main Results}

\begin{table*}[!t]
    \centering
    \caption{\textbf{Main results on COCO \texttt{val2017}.} We mainly compare \model{} with recent query-based methods on COCO \texttt{val2017}. The inference speeds are measured on a single NVIDIA 2080Ti GPU.}
    \resizebox{\textwidth}{!}{
    \begin{tabular}{l|l|l|c|ccc|ccc|c}
    Method  & Backbone & Feature & Epochs & AP & AP$_{50}$ & AP$_{75}$  & AP$_{S}$ & AP$_{M}$  & AP$_{L}$ & FPS \\
    \tline
    RetinaNet~\cite{FocalLinGGHD17} & ResNet-50 & FPN  & $36$  & $38.7$  & $58.0$  & $41.5$  & $23.3$  & $42.3$  & $50.3$  & $20.2$ \\
    Faster R-CNN~\cite{ren2015faster} &  ResNet-50 & FPN & $36$ & $40.2$ & $61.0$ & $43.8$ & $24.2$ & $43.5$ & $52.0$ & $22.1$ \\
    Cascade R-CNN~\cite{CascdeCaiV21} &  ResNet-50 & FPN & $36$ & $43.4$ & $61.3$ & $47.3$ & $25.9$ & $46.8$ & $56.8$ & $17.5$ \\
    DETR~\cite{DETRCarionMSUKZ20} &  ResNet-50  & Encoder & $500$ & $42.0$ & $62.4$ & $44.2$ & $20.5$ & $45.8$ & $61.1$ & $23.7$ \\
    Deformable DETR~\cite{DeformableZhu20} &  ResNet-50 & DeformEncoder & $50$ & $43.8$ & $62.6$ & $47.7$ & $26.4$ & $47.1$ & $58.0$ & $12.6$ \\
    Efficient DETR~\cite{EfficientYao21} &  ResNet-50 & DeformEncoder  & $36$    & $44.2$ & $62.2$ & $48.0$ & $28.4$ & $47.5$ & $56.6$ & - \\
    Sparse R-CNN (6 Stages, 100 Queries)~\cite{sun2020sparse} &  ResNet-18   & FPN   & $36$    & $38.5$  & $55.9$  & $41.8$ & $22.8$   & $40.2$ & $50.4$ & $29.9$ \\
    Sparse R-CNN (6 Stages, 100 Queries)~\cite{sun2020sparse}  &  ResNet-50  & FPN   & $36$    & $42.3$  & $61.2$  & $45.7$ & $26.7$   & $44.6$ & $57.6$ & $20.9$ \\
    Sparse R-CNN (6 Stages, 300 Queries)~\cite{sun2020sparse} &  ResNet-50   & FPN   & $36$    & $44.5$  & $63.4$  & $48.2$ & $26.9$   & $47.2$ & $59.5$ & $20.6$ \\
    \tline
    \model~($100$ Queries)   &  ResNet-18      & FPN & $36$ & $36.4$ & $53.8$ & $39.7$ & $23.0$ & $38.3$ & $46.1$ & $45.4$ \\
    \cascademodel~($100$ Queries)   &  ResNet-18      & FPN & $36$ & $38.5$ & $56.1$ & $41.9$ & $23.4$ & $40.4$ & $49.5$ & $37.3$ \\
    \hline
    \model~($100$ Queries)       &  ResNet-50 & FPN & $36$ & $41.3$ & $59.4$ & $44.9$ & $26.7$ & $44.2$ & $52.4$ & $26.0$ \\
    \cascademodel~($100$ Queries) & ResNet-50 & FPN   & $36$ & $43.0$ & $61.3$  & $46.8$ & $28.3$   & $45.7$    & $55.5$ & $24.5$ \\
    \cascademodel~($300$ Queries) & ResNet-50 & FPN   & $36$ & $44.6$ & $63.1$  & $48.9$ & $29.5$   & $47.4$    & $57.5$ & $24.3$\\
    \tline
    RetinaNet~\cite{FocalLinGGHD17} &  ResNet-101 & FPN & $36$ & $40.4$ & $60.2$ & $43.2$ & $24.0$ & $44.3$ & $52.2$ & $15.2$ \\
    Faster R-CNN~\cite{ren2015faster} &  ResNet-101 &  FPN & $36$ & $42.0$ & $62.5$ & $45.9$ & $25.2$ & $45.6$ & $54.6$ & $16.0$ \\
    DETR~\cite{DETRCarionMSUKZ20} &  ResNet-101 & Encoder & $500$ & $43.5$ & $63.8$ & $46.4$ & $21.9$ & $48.0$ & $61.8$ & $17.2$ \\
    DETR-DC5~\cite{DETRCarionMSUKZ20} &  ResNet-101 & Encoder & $500$ & $44.9$ & $64.7$ & $47.7$ & $23.7$ & $49.5$ & $62.3$ & $8.7$ \\
    TSP-RCNN~\cite{RethinkingSun20} &  ResNet-101 & FPN & $36$ & $44.8$ & $63.8$ & $49.2$ & $29.0$ & $47.9$ & $57.1$ & 9* \\
    Efficient DETR~\cite{EfficientYao21} &  ResNet-101 & DeformEncoder & $36$ & $45.2$ & $63.7$ & $48.8$ & $28.8$ & $49.1$ & $59.0$ & - \\
    Sparse R-CNN ($6$ Stages, $300$ Queries)~\cite{sun2020sparse}    &  ResNet-101 & FPN   & $36$    & $45.6$  & $64.6$  & $49.5$ & $28.3$   & $48.3$ & $61.6$ & $15.2$ \\
    \tline
    \cascademodel~($100$ Queries) &  ResNet-101  & FPN & $36$ & $43.9$ & $62.2$ & $47.9$ & $26.7$ & $46.7$ & $58.1$ & $17.7$ \\
    \cascademodel~($300$ Queries) &  ResNet-101 & FPN & $36$ & $45.8$ & $64.4$ & $49.9$ & $30.1$ & $48.5$ & $60.1$ & $17.4$ \\
    \end{tabular}
    }
    \label{tab: main_results_coco}
    \vspace{-7pt}
\end{table*}

We give a full comparisons with main stream detectors~\cite{FocalLinGGHD17, ren2015faster, CascdeCaiV21} and recently proposed query-based or transformer-based methods. Tab.~\ref{tab: main_results_coco} and Tab.~\ref{tab: coco_test} show the results of the proposed \model{} on COCO \texttt{val2017} and \texttt{test-dev}, respectively. 
Using one detector head and $100$ queries, \model{} obtains a higher AP with faster inference speed than typical one-stage methods RetinaNet and two-stage methods Faster R-CNN. Under similar AP results, our \model{} with ResNet-50 is $1.2\times$ faster than Faster R-CNN with same backbone. Cascade R-CNN is a popular two-stage high-performance detector integrating iterative structure heads. By using only 2-stage cascade head and $300$ queries, \cascademodel{} surpasses Cascade R-CNN $1.1$ AP  while runs with higher FPS.

Compared to the popular query-based or transformer-based methods, our method also has competitive results considering AP and inference speed. As shown in Tab.~\ref{tab: main_results_coco}, we use a shorter schedule than ~\cite{DETRCarionMSUKZ20, DeformableZhu20} and fewer stages of decoders than \cite{DETRCarionMSUKZ20, DeformableZhu20,RethinkingSun20}. Besides, different from ~\cite{EfficientYao21}, we don't rely on powerful one-stage detector and transformer-encoder. We obtain $44.6$ AP with ResNet-50 and $45.8$ AP with ResNet-101 when using $300$ queries, which are better than most query-based methods.

\subsection{Ablation Experiments}
\label{sec: ab_study}

\paragraph{The recall of Query Generation Networks}

\begin{figure*}[]
    \centering
    \subfigure[]{
    \begin{minipage}{0.3\linewidth}
        \includegraphics[width=\linewidth]{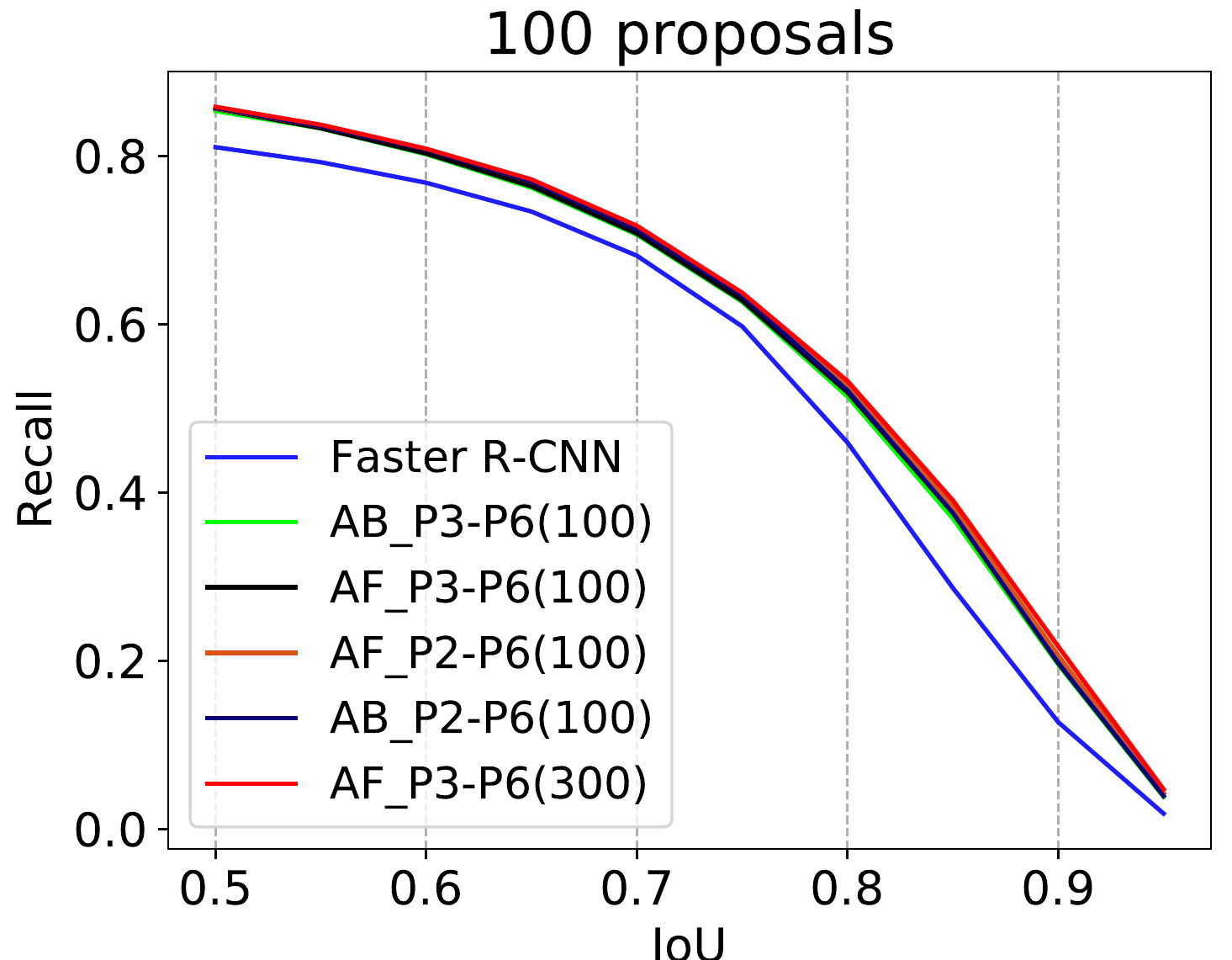}
    \end{minipage}}
    \subfigure[]{
    \begin{minipage}{0.3\linewidth}
        \includegraphics[width=\linewidth]{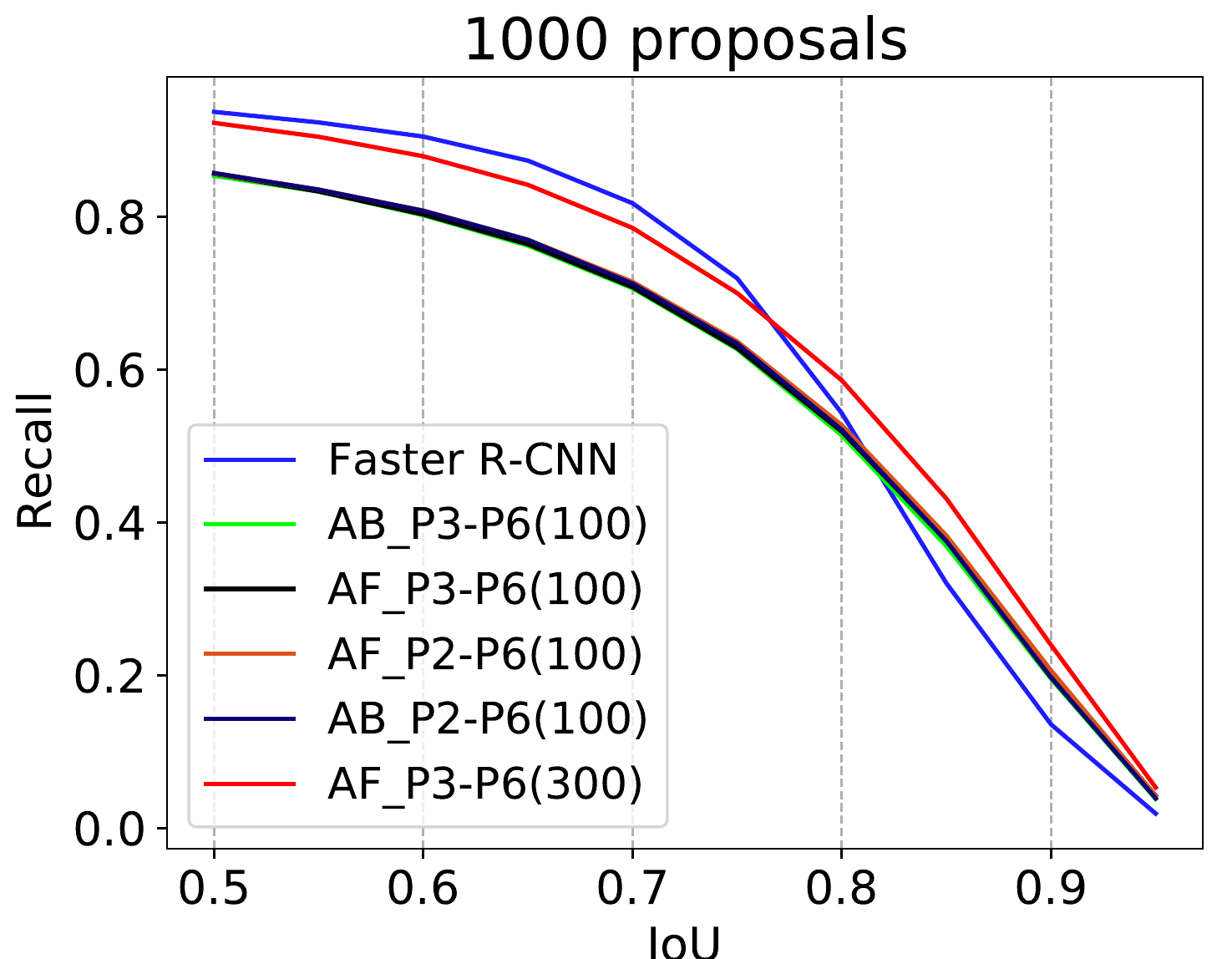}
    \end{minipage}}
    \begin{minipage}[!t]{0.31\linewidth}
        \begingroup
        \fontsize{8pt}{-2pt}\selectfont
        (c) \textbf{AR$_{100}$ and AR$_{1000}$ comparison.}\\
        Our QGN have a higher AR$_{100}$ using only 100 proposals and higher AR$_{1000}$ using 300 proposals than RPN.
        \endgroup
        \vspace{2pt}
        \scalebox{0.8}{
            \begin{tabular}{c|c|c}
             & AR$_{100}$ & AR$_{1000}$ \\
            \tline
            RPN                   & 52.75     & 61.93 \\
            Ours($p=100$)        & 57.31     & 57.31 \\
            Ours($p=300$)        & 58.17     & 63.42 \\
            \end{tabular}}
    \end{minipage}
    \vspace{-10pt}
    \caption{\textbf{Recall \textit{v.s.} IoU on COCO 2017 validation set.} We use ResNet50-FPN as our backbone. $AB$, $AF$ denotes anchor-based, anchor-free respectively. $Px$-$Py$ represents using features from level $x$ to level $y$. The P2 level feature brings no gains to our overall recall. Our QGN obtains higher recall using fewer proposals.}
    \label{fig: recall_iou}
\end{figure*}

Can our proposed Query Generation Networks (QGN) generate high-recall proposals? To answer this question, we perform various settings of QGN to compare recall with popular RPN of ~\cite{ren2015faster}. For fair comparison, we simply replace the RPN in ~\cite{ren2015faster} with QGN, and the strategies and structures of other parts remain the same.
In Fig~\ref{fig: recall_iou}, we compare the recall of proposals in different IoU threshold. When using 100 proposals, our method performs better than Faster R-CNN. As we rise the proposals of Faster R-CNN to 1000, our method still have advantages with high IoU as shown in Fig~\ref{fig: recall_iou}(b). 
With top-300 proposals, the QGN achieves better AR$_{1000}$ compared to Faster R-CNN. 
Besides, we observe little difference in anchor-based and anchor-free methods. 
The P2 feature of backbone which has $\frac{1}{4}\times$ resolution \wrt~ the input image is not used in our method considering efficiency.

\begin{table*}[ht]
    \centering
    \caption{\textbf{Main results on COCO \texttt{test-dev}}. We compare \model{} with the popular one-stage and two-stage detectors on COCO \texttt{test-dev}.}
    \resizebox{0.98\textwidth}{!}{
    \begin{tabular}{l|l|ccc|ccc}
    Method  & Backbone  & AP & AP$_{50}$ & AP$_{75}$  & AP$_{S}$ & AP$_{M}$  & AP$_{L}$\\
    \tline
    RetinaNet~\cite{FocalLinGGHD17} & ResNet-101   & $39.1$  & $59.1$  & $42.3$  & $21.8$  & $42.7$  & $50.2$ \\
    FSAF~\cite{zhu2019feature}  & ResNet-101  & $40.9$ & $61.5$ & $44.0$ & $24.0$ & $44.2$ & $51.3$ \\
    FCOS~\cite{tian2019fcos} & ResNet-101   & $41.5$ & $60.7$ & $45.0$ & $24.4$ & $44.8$ & $51.6$ \\
    RepPoints~\cite{yang2019reppoints} & ResNet-101-DCN   & $45.0$ & $66.1$ & $49.0$ & $26.6$ & $48.6$ & $57.5$ \\
    ATSS~\cite{zhang2020bridging} & ResNet-101  & $43.6$ & $62.1$ & $47.4$ & $26.1$ & $47.0$ & $53.6$ \\
    ATSS~\cite{zhang2020bridging} & ResNet-101-DCN  & $46.3$ & $64.7$ & $50.4$ & $27.7$ & $49.8$ & $58.4$ \\
    \hline
    Faster RCNN~\cite{ren2015faster} & ResNet-101  & $36.2$ & $59.1$ & $39.0$ & $18.2$ & $39.0$ & $48.2$ \\
    Libra RCNN~\cite{pang2019libra} & ResNet-101   & $41.1$ & $62.1$ & $44.7$ & $23.4$ & $43.7$ & $52.5$ \\
    Cascade R-CNN~\cite{CascdeCaiV21} & ResNet-101  & $42.8$ & $62.1$ & $46.3$ & $23.7$ & $45.5$ & $55.2$ \\
    Dynamic RCNN~\cite{zhang2020dynamic} & ResNet-101   & $44.7$ & $63.6$ & $49.1$ & $26.0$ & $47.4$ & $57.2$ \\
    TSP-RCNN~\cite{RethinkingSun20}  & ResNet-101   & $46.6$ & $66.2$ & $51.3$ & $28.4$ & $49.0$ & $58.5$ \\
    Sparse RCNN ($6$ Stages, $300$ Queries)~\cite{sun2020sparse}  & ResNeXt-101   & $46.9$ & $66.3$ & $51.2$ & $28.6$ & $49.2$ & $58.7$ \\
    \tline
    \cascademodel~($100$ Queries) & ResNet-101    & $44.4$ & $63.3$ & $48.3$ & $26.4$ & $46.6$ & $55.5$ \\
    \cascademodel~($300$ Queries) & ResNet-101    & $46.0$ & $64.9$ & $50.3$ & $28.5$ & $48.6$ & $56.9$ \\
    \cascademodel~($300$ Queries) & ResNeXt-101    & $47.0$ & $66.4$ & $51.5$ & $29.5$ & $49.4$ & $57.7$ \\
    \end{tabular}
    }
    \label{tab: coco_test}
    \vspace{-3pt}
\end{table*}

\paragraph{Effects of the components}
Tab.~\ref{tab:comparision_component}(a) shows the results of adding an extra RoI-level self-attention and using P7 features on the single R-CNN head or the cascade head respectively. Due to the hardly-optimized characteristic of interaction module indicated in ~\cite{RethinkingSun20}, our method with isolated head have a relatively poor capability extracting object embedding and perform worse than ~\cite{ren2015faster}. We additionally add a self-attention module to enhance the object features in RoI-feature levels and an interesting phenomenon appeared. There is $0.9$ AP gains in single detector-head but only $0.2$ AP gains in cascade head scenario. It proves that the interactive module has a stronger ability to extract object feature when using cascade head. Our RoI-level self-attention module can alleviate this problem effectively. Besides, using P3-P7 features for QGN is better than P3-P6 features.

\begin{table}
\caption{\small\textbf{Ablation studies.} We provide the ablation studies about (a) the components in \model{}, (b) query generation paradigm in QGN, (c) the generation of featurized queries, (d) position encoding, and (d) the generalization comparison with Sparse R-CNN on CrowdHuman.}
\vspace{1pt}
\begin{minipage}{\linewidth}
    \begingroup
    \small{(a) \textbf{The effect of different component.} When using 1-stage R-CNN head, the RoI-level self-attention brings 0.9 AP gains. At this time, we get the similar results to Faster R-CNN-R50.}
    \endgroup
    \vskip 0.05 in
    \centering
    \renewcommand\arraystretch{1.1}
    \scalebox{0.8}{
        \begin{tabular}{l|cc|ccc|c}
     & Self-Attn. & w/ P7 & \aps & Time(ms) \\
    \tline
    \multirow{3}*{\model{}}      &  &  & $39.2$ & $56.9$ & $43.0$ & $37.3$ \\
                      & \checkmark &  & $40.1$ & $58.2$ & $44.0$ & $37.6$ \\
            & \checkmark & \checkmark & $41.3$ & $59.4$ & $44.9$ & $38.4$ \\
    \tline
    \multirow{3}*{Cascade \model{}}      &  &  & $42.1$ & $60.3$ & $45.9$ & $39.8$ \\
                      & \checkmark &  & $42.3$ & $60.5$ & $46.0$ & $40.2$ \\
            & \checkmark & \checkmark & $42.8$ & $61.1$ & $46.9$ &  $40.7$ \\
    \end{tabular}}
    \label{tab:comparision_component}
\end{minipage}
\vskip 0.1 in 
\begin{minipage}[t]{0.49\linewidth}
\begingroup
\small{(b) \textbf{Query generation paradigm of QGN.} We evaluate the performance using anchor-based or anchor-free paradigms in QGN. The anchor-free paradigm provides better results.}
\endgroup
\vskip 0.05 in
\centering
\renewcommand\arraystretch{1.1}
\scalebox{0.80}{
    \begin{tabular}{l|ccc|c}
     & \aps & Time (ms) \\
    \tline
    anchor-based & $38.2$ & $55.3$ & $41.4$ & $39.1$ \\
    anchor-free  & $39.2$ & $56.6$ & $25.1$ & $37.3$ \\
    \end{tabular}}
\label{tab:sample_metrics}
\end{minipage}
\hfill
\begin{minipage}[t]{.49\linewidth}
\begingroup
\footnotesize{ (c) \textbf{Featurized query generation.} We evaluate different settings for generating featurized queries.}
\endgroup
\vskip 0.05 in
\centering
\renewcommand\arraystretch{1.1}
\scalebox{0.80}{
\begin{tabular}{l|ccc}
    Settings & \aps \\
    \tline
    Learnt embedding        & $33.9$ & $48.7$ & $37.2$ \\
    $1\times1$ conv.    & $39.2$ & $56.9$ & $43.0$ \\
    $3\times3$ conv.    & $39.2$ & $56.6$ & $43.0$ \\
    Stacked conv.    & $39.4$ & $57.3$ & $43.0$ \\
    \end{tabular}}
\label{tab:feat_query}
\end{minipage}
\vskip 0.1 in
\begin{minipage}[t]{0.49\linewidth}
\begingroup
\footnotesize{ (d) \textbf{Position encoding.} We evaluate the effect of adding position encoding for query features and RoI features, which shows minor improvements.}
\endgroup
\vskip 0.05 in
\centering
\renewcommand\arraystretch{1.1}
\scalebox{0.8}{
        \begin{tabular}{cc|ccc}
        QGN & R-CNN & \aps \\
        \tline
        None & None          & $42.8$ & $61.0$ & $46.9$ \\
        Sine at input & None & $42.8$ & $61.1$ & $46.9$ \\
        None & Sine at input & $42.7$ & $61.2$ & $46.5$ \\
        Sine at input & Sine at input & $42.7$ & $61.1$ & $46.4$\\
    \end{tabular}}
\label{tab:pos_embed}
\end{minipage}
\hfill
\begin{minipage}[t]{0.49\linewidth}
\begingroup
\footnotesize{ (e) \textbf{Generalization capability.} We directly apply the COCO-pretrained models to evaluate on CrowdHuman \textit{val} without
fine-tuning.}
\endgroup
\vskip 0.05 in
\centering
\renewcommand\arraystretch{1.1}
\scalebox{0.8}{
    \begin{tabular}{c|l|cl}
    Queries & Methods  & COCO AP & AP$_{50}$ \\
    \tline
    \multirow{2}*{100}  & Sparse R-CNN & $42.8$ & $40.0$ \\
                                & Ours         & $43.0$ & $42.8$ $^{\textcolor{red}{+2.8}}$ \\
    \tline
    \multirow{2}*{300}  & Sparse R-CNN & $45.0$ & $44.5$ \\
                                & Ours         & $44.6$ & $46.6^{\textcolor{red}{+2.1}}$ \\
    \end{tabular}}
\label{tab:generalization_capability}
\end{minipage}
\vspace{-10pt}
\end{table}

\paragraph{Query generation paradigm of QGN}
We compare the two typical generation paradigm in \model{}, \ie, anchor-based paradigm and anchor-free paradigm.
As for the anchor-based paradigm, we follow \cite{ren2015faster} and define 9 anchors per location.
While for anchor-free head, each location corresponds to only one object.
For both anchor-free or anchor-based paradigm, we adopt the same training loss as discussed in Sec.~\ref{qgn_loss}
Tab.~\ref{tab:sample_metrics}(b) shows that anchor-free head achieves better accuracy.
Compared to anchor-free paradigm, anchor-based paradigm tends to output multiple query boxes per location but shares the pixel features, which leads to ambiguity for refinement in R-CNN head, \ie, multiple query boxes share the same query features.

\paragraph{Generation of featurized query}
We compare different settings of featurized query branch in Tab.~\ref{tab:feat_query}(c). When the queries are learnable weight same as~\cite{DETRCarionMSUKZ20, sun2020sparse}, the AP drop to $34.0$. This result shows that the one-to-one correspondence between boxes and featurized queries is necessary. Using a $3\times3$ conv layer to generate the query has sight difference between using $3\times3$ ones. Besides, when we stack two layers (one $3\times3$ conv following a $1\times1$ conv) to get the featurized queries, there are 0.2 AP gains compared to basic ones. For simplicity, we use a single $1\times 1$ conv layer by defaults.

\begin{wrapfigure}[20]{r}{0.50\linewidth}
\centering
\vspace{-10pt}
\includegraphics[width=\linewidth]{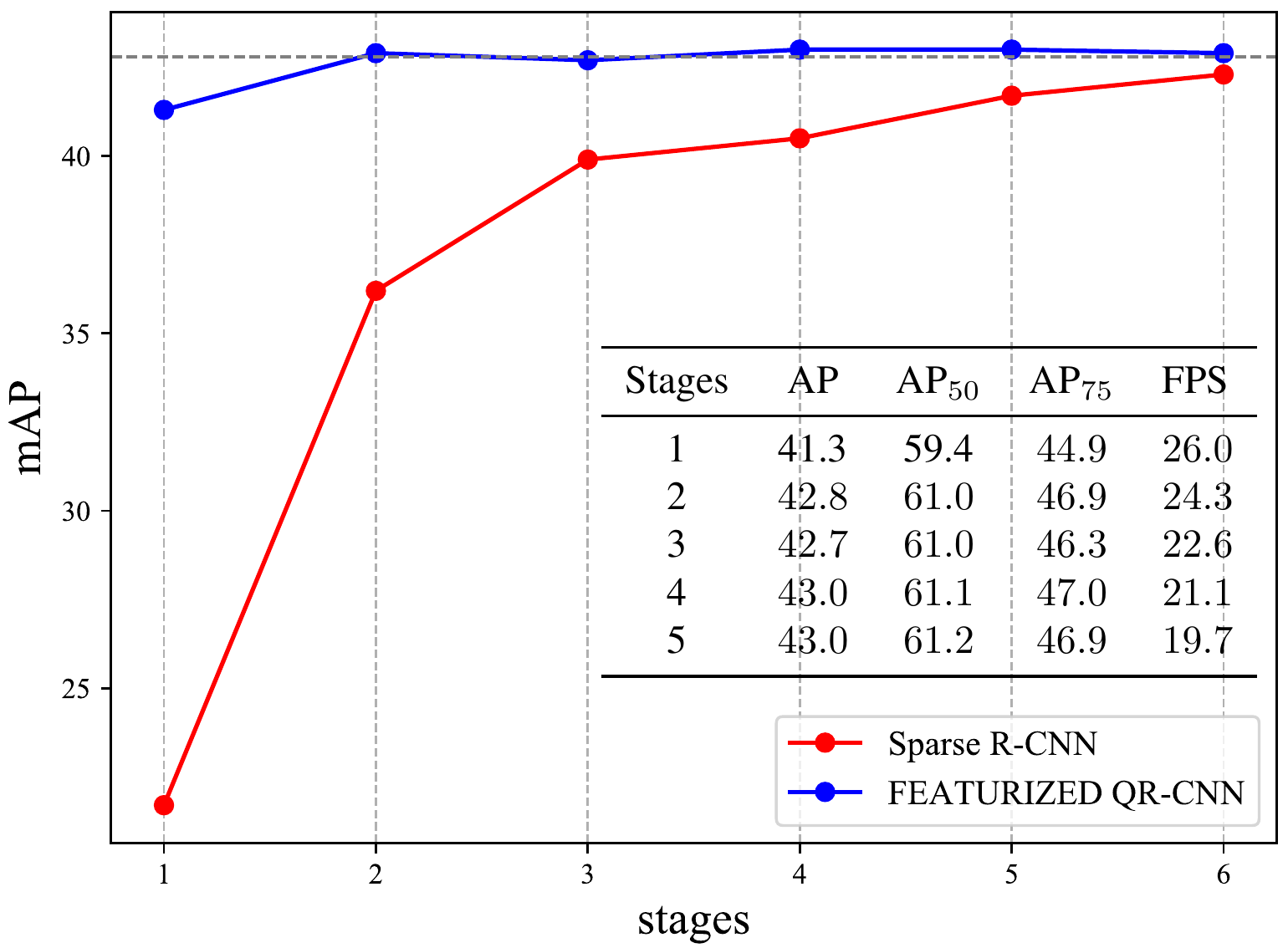}
\vspace{-20pt}
\caption{\textbf{Effect of number of stages in Cascade Head.} We evaluate the effect of different number of stages. \model~with cascade head outperforms Sparse R-CNN with 6 stages. Using QGN provides better initial queries and only requires 1 $\sim$ 2 stages for refinements, and adding more stages shows minor improvement but incurs computation cost.}
\label{fig: stage_ap}
\end{wrapfigure}

\paragraph{Effect of position encoding}
Position encodings act as the key component in transformer-based methods~\cite{DETRCarionMSUKZ20,DeformableZhu20,conditionaldetrMengCFZLYS021}, we also incorporate spatial position encodings into query features in QGN or region features in R-CNN.
Tab.~\ref{tab:pos_embed}(d) shows that using position encodings has no impact on our method.

\paragraph{Generalization ability of queries} In order to evaluate the generalization ability of our proposed featurized query, we perform experiments on CrowdHuman dataset~\cite{shao2018crowdhuman}. Intuitively, we directly apply the COCO-pretrained models to evaluate on \textit{val} set without fine-tuning. Tab.~\ref{tab:generalization_capability}(e) shows that our method significantly outperforms Sparse R-CNN on CrowdHuman. In the case where the COCO-AP of two methods are close, our method outperforms Sparse R-CNN by nearly $2$ AP, showing better generalization ability.

\paragraph{Effect of numbers of stages in Cascade Head}
\label{sec:num_deocders}
Our method alleviates the dependence of the query-based method on the iterative structures. In Fig.~\ref{fig: stage_ap}, we evaluate the impact of the iterative structures on our method and Sparse R-CNN. For our method, when the number of stages in cascade head exceeds three, the performance gain is almost negligible. We use two-stage cascade head by default, which is more efficient than Sparse R-CNN but provides competitive accuracy.

\paragraph{Visualization and Timing} 

\begin{figure}
\centering
\vspace{-5pt}
\includegraphics[width={0.95\textwidth}]{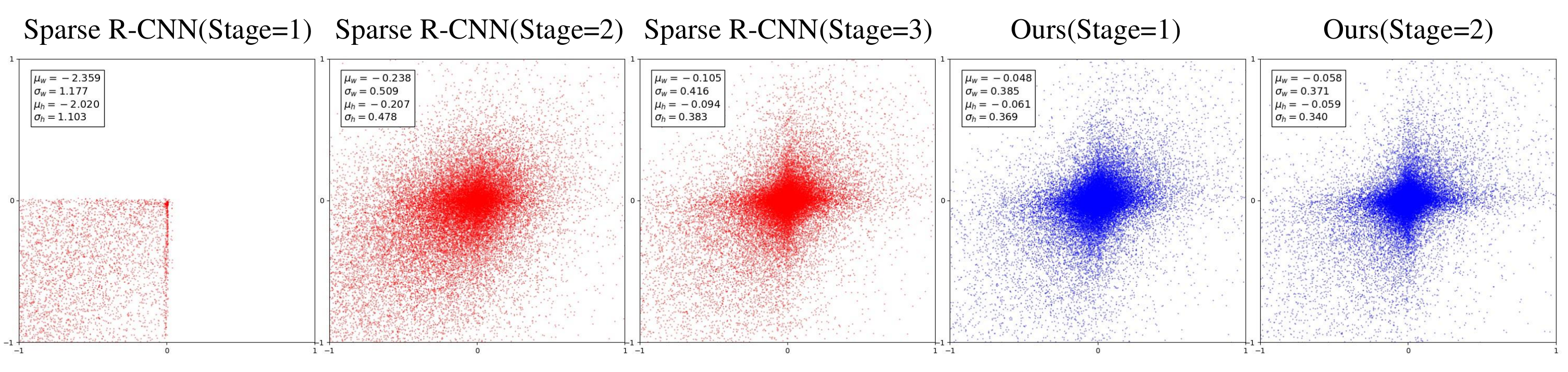}
\caption{\textbf{Distribution of the distance vector of ($\delta_w$, $\delta_h$).} The proposals of QGN have smaller distance with GT than the outputs of Sparse R-CNN at the same stage.}
\label{fig: proposals_distribution}
\end{figure}

We visualize the distribution of $\Delta$=($\delta_w$, $\delta_h$) in Fig.~\ref{fig: proposals_distribution}, where,
\begin{equation}
    {\delta_w} = (g_x - b_x)/b_w, {\delta_h} = (g_y - b_y)/b_h,
\end{equation}
The R-CNN head performs adjustments on proposals $b$=($b_x$, $b_y$, $b_w$, $b_h$), and the proposals gradually approach the GT $g$=($b_x$, $b_y$, $b_w$, $b_h$). For the first stage of R-CNN, the proposals are generated from QGN. As for the later stages, the proposals are from previous stage. This distribution reveals the convergence of region proposals. Our method has better convergence properties than Sparse R-CNN when using fewer stages. Besides, we visualize the bounding boxes of different stages in Fig.~\ref{fig: rpn_proposals}. The query boxes generated by QGN can cover most of the instances in the images. The R-CNN head can refine the proposals progressively and get accurate predictions at last.

We report the inference latency of different module in \model{}. The results are shown in Tab.\ref{tab:timming}. 
We divide the query-based detectors into four parts, namely, backbone, encoder, query generation, decoder. 
Compared with with other popular query-based detector, our method uses a lightweight QGN to generate queries and do not rely on transformer encoders,
and the decoders are much efficient.
Tab.~\ref{tab:timming} shows that \model{} achieves better trade-off.

\begin{table}[h]
    \centering
    \caption{\textbf{Latency (ms) and AP comparison.} E and D denote encoder and decoder respectively.}
    \resizebox{1.\textwidth}{!}{
    \begin{tabular}{l|cccc|cc}
    Method & Backbone & Encoder & QueryGene. & Decoder & Total(ms) & AP \\
    \tline
    Deform-DETR (300 Queries, 6E, 6D) & 29.5 & 33.9  & 0.2   & 15.4 & 79.3 & 43.8 \\
    Efficient-DETR (300 Queries, 3E, 1D) & 29.5 & 18.1  & 4.8   & 2.9 & 55.3 & 44.2 \\
    Sparse R-CNN (100 Queries, 6D) & 28.6 & -   & 0.3   & 17.6 & 47.8 & 42.3 \\
    Sparse R-CNN (300 Queries, 6D) & 28.6 & -   & 0.3   & 18.3 & 48.8 & 44.5\\
    \tline
    \model{} (100 Queries)   & 29.8 & -   & 3.3   & 3.8  & 38.4 & 41.3 \\
    \cascademodel (100 Queries)   & 29.8 & -   & 3.3   & 6.7  & 40.3 & 42.8 \\
    \cascademodel (300 Queries)   & 29.8 & -   & 3.3   & 7.1  & 41.2 & 44.6 \\
    \end{tabular}}
    \vspace{-1pt}
    \label{tab:timming}
\end{table}

\section{Conclusion}
In this paper, we propose \model{}, aims to introduce query-based principle into R-CNN based detector while alleviate the dependence of iterative decoders in Sparse R-CNN. Specifically, we propose a query generation network (QGN) to generate image-aware object queries. By sharing features from backbone, the object query can get instance-level context in a nearly cost-free way. Further, we adopt one region-based decoder to refine the object proposals. Extensive experiments have demonstrated the effectiveness of our proposed methods. We believe that \model{} can serve as a new and standard tool for object detection due to its simplicity and strong performance.

\bibliographystyle{unsrt}
\bibliography{ref}

\begin{thebibliography}{10}

\bibitem{girshick2015fast}
Ross Girshick.
\newblock Fast r-cnn.
\newblock In {\em Proceedings of the IEEE international conference on computer
  vision}, pages 1440--1448, 2015.

\bibitem{ren2015faster}
Shaoqing Ren, Kaiming He, Ross Girshick, and Jian Sun.
\newblock Faster r-cnn: Towards real-time object detection with region proposal
  networks.
\newblock In {\em Advances in neural information processing systems}, pages
  91--99, 2015.

\bibitem{COCO}
Tsung{-}Yi Lin, Michael Maire, Serge~J. Belongie, James Hays, Pietro Perona,
  Deva Ramanan, Piotr Doll{\'{a}}r, and C.~Lawrence Zitnick.
\newblock Microsoft {COCO:} common objects in context.
\newblock In {\em ECCV}, 2014.

\bibitem{sun2020sparse}
Peize Sun, Rufeng Zhang, Yi~Jiang, Tao Kong, Chenfeng Xu, Wei Zhan, Masayoshi
  Tomizuka, Lei Li, Zehuan Yuan, Changhu Wang, et~al.
\newblock Sparse r-cnn: End-to-end object detection with learnable proposals.
\newblock In {\em CVPR}, 2021.

\bibitem{he2017mask}
Kaiming He, Georgia Gkioxari, Piotr Doll{\'a}r, and Ross Girshick.
\newblock Mask r-cnn.
\newblock In {\em ICCV}, 2017.

\bibitem{CascdeCaiV21}
Zhaowei Cai and Nuno Vasconcelos.
\newblock Cascade {R-CNN:} high quality object detection and instance
  segmentation.
\newblock {\em {IEEE} Trans. Pattern Anal. Mach. Intell.}, 2021.

\bibitem{PichGirshick14}
Ross Girshick, Jeff Donahue, Trevor Darrell, and Jitendra Malik.
\newblock Rich feature hierarchies for accurate object detection and semantic
  segmentation.
\newblock In {\em CVPR}, 2014.

\bibitem{selective_search}
Jasper R.~R. Uijlings, Koen E.~A. van~de Sande, Theo Gevers, and Arnold W.~M.
  Smeulders.
\newblock Selective search for object recognition.
\newblock {\em {IJCV}}, 2013.

\bibitem{chen2019hybrid}
Kai Chen, Jiangmiao Pang, Jiaqi Wang, Yu~Xiong, Xiaoxiao Li, Shuyang Sun,
  Wansen Feng, Ziwei Liu, Jianping Shi, Wanli Ouyang, et~al.
\newblock Hybrid task cascade for instance segmentation.
\newblock In {\em CVPR}, 2019.

\bibitem{guided_anchor}
Jiaqi Wang, Kai Chen, Shuo Yang, Chen~Change Loy, and Dahua Lin.
\newblock Region proposal by guided anchoring.
\newblock In {\em CVPR}, 2019.

\bibitem{cascade_rpnuJPY19}
Thang Vu, Hyunjun Jang, Trung~X. Pham, and Chang~Dong Yoo.
\newblock Cascade {RPN:} delving into high-quality region proposal network with
  adaptive convolution.
\newblock In {\em {NeurIPS}}, 2019.

\bibitem{DETRCarionMSUKZ20}
Nicolas Carion, Francisco Massa, Gabriel Synnaeve, Nicolas Usunier, Alexander
  Kirillov, and Sergey Zagoruyko.
\newblock End-to-end object detection with transformers.
\newblock In {\em ECCV}, 2020.

\bibitem{DeformableZhu20}
Xizhou Zhu, Weijie Su, Lewei Lu, Bin Li, Xiaogang Wang, and Jifeng Dai.
\newblock Deformable detr: Deformable transformers for end-to-end object
  detection.
\newblock {\em ArXiv:2010.04159}, 2020.

\bibitem{dai2017deformable}
Jifeng Dai, Haozhi Qi, Yuwen Xiong, Yi~Li, Guodong Zhang, Han Hu, and Yichen
  Wei.
\newblock Deformable convolutional networks.
\newblock In {\em ICCV}, 2017.

\bibitem{fang2021you}
Yuxin Fang, Bencheng Liao, Xinggang Wang, Jiemin Fang, Jiyang Qi, Rui Wu,
  Jianwei Niu, and Wenyu Liu.
\newblock You only look at one sequence: Rethinking transformer in vision
  through object detection.
\newblock In {\em NeurIPS}, 2021.

\bibitem{gao2021fast}
Peng Gao, Minghang Zheng, Xiaogang Wang, Jifeng Dai, and Hongsheng Li.
\newblock Fast convergence of detr with spatially modulated co-attention.
\newblock {\em ICCV}, 2021.

\bibitem{conditionaldetrMengCFZLYS021}
Depu Meng, Xiaokang Chen, Zejia Fan, Gang Zeng, Houqiang Li, Yuhui Yuan, Lei
  Sun, and Jingdong Wang.
\newblock Conditional {DETR} for fast training convergence.
\newblock In {\em {ICCV}}, 2021.

\bibitem{wang2021anchor}
Yingming Wang, Xiangyu Zhang, Tong Yang, and Jian Sun.
\newblock Anchor detr: Query design for transformer-based detector.
\newblock {\em arXiv preprint arXiv:2109.07107}, 2021.

\bibitem{dabdetr}
Shilong Liu, Feng Li, Hao Zhang, Xiao Yang, Xianbiao Qi, Hang Su, Jun Zhu, and
  Lei Zhang.
\newblock {DAB-DETR:} dynamic anchor boxes are better queries for {DETR}.
\newblock In {\em ICLR}, 2022.

\bibitem{EfficientYao21}
Zhuyu Yao, Jiangbo Ai, Boxun Li, and Chi Zhang.
\newblock Efficient detr: Improving end-to-end object detector with dense
  prior.
\newblock {\em ArXiv:2104.01318}, 2021.

\bibitem{roh2021sparse}
Byungseok Roh, JaeWoong Shin, Wuhyun Shin, and Saehoon Kim.
\newblock Sparse detr: Efficient end-to-end object detection with learnable
  sparsity.
\newblock In {\em ICLR}, 2022.

\bibitem{fang2021instances}
Yuxin Fang, Shusheng Yang, Xinggang Wang, Yu~Li, Chen Fang, Ying Shan, Bin
  Feng, and Wenyu Liu.
\newblock Instances as queries.
\newblock In {\em ICCV}, 2021.

\bibitem{stewart2016end}
Russell Stewart, Mykhaylo Andriluka, and Andrew~Y Ng.
\newblock End-to-end people detection in crowded scenes.
\newblock In {\em CVPR}, 2016.

\bibitem{DBLP:conf/cvpr/LinDGHHB17}
Tsung{-}Yi Lin, Piotr Doll{\'{a}}r, Ross~B. Girshick, Kaiming He, Bharath
  Hariharan, and Serge~J. Belongie.
\newblock Feature pyramid networks for object detection.
\newblock In {\em CVPR}, 2017.

\bibitem{EndWang20}
Jianfeng Wang, Lin Song, Zeming Li, Hongbin Sun, Jian Sun, and Nanning Zheng.
\newblock End-to-end object detection with fully convolutional network.
\newblock In {\em CVPR}, 2021.

\bibitem{tian2019fcos}
Zhi Tian, Chunhua Shen, Hao Chen, and Tong He.
\newblock Fcos: Fully convolutional one-stage object detection.
\newblock In {\em ICCV}, 2019.

\bibitem{AttentionVaswani17}
Ashish Vaswani, Noam Shazeer, Niki Parmar, Jakob Uszkoreit, Llion Jones,
  Aidan~N Gomez, Lukasz Kaiser, and Illia Polosukhin.
\newblock Attention is all you need.
\newblock {\em ArXiv:1706.03762}, 2017.

\bibitem{wu2019detectron2}
Yuxin Wu, Alexander Kirillov, Francisco Massa, Wan-Yen Lo, and Ross Girshick.
\newblock Detectron2.
\newblock \url{https://github.com/facebookresearch/detectron2}, 2019.

\bibitem{loshchilov2017decoupled}
Ilya Loshchilov and Frank Hutter.
\newblock Decoupled weight decay regularization.
\newblock {\em ArXiv:1711.05101}, 2017.

\bibitem{imagenet}
Jia Deng, Wei Dong, Richard Socher, Li{-}Jia Li, Kai Li, and Fei{-}Fei Li.
\newblock Imagenet: {A} large-scale hierarchical image database.
\newblock In {\em CVPR}, 2009.

\bibitem{he2016deep}
Kaiming He, Xiangyu Zhang, Shaoqing Ren, and Jian Sun.
\newblock Deep residual learning for image recognition.
\newblock In {\em CVPR}, 2016.

\bibitem{FocalLinGGHD17}
Tsung{-}Yi Lin, Priya Goyal, Ross~B. Girshick, Kaiming He, and Piotr
  Doll{\'{a}}r.
\newblock Focal loss for dense object detection.
\newblock In {\em {ICCV}}, 2017.

\bibitem{RethinkingSun20}
Zhiqing Sun, Shengcao Cao, Yiming Yang, and Kris Kitani.
\newblock Rethinking transformer-based set prediction for object detection.
\newblock {\em ICCV}, 2021.

\bibitem{zhu2019feature}
Chenchen Zhu, Yihui He, and Marios Savvides.
\newblock Feature selective anchor-free module for single-shot object
  detection.
\newblock In {\em CVPR}, 2019.

\bibitem{yang2019reppoints}
Ze~Yang, Shaohui Liu, Han Hu, Liwei Wang, and Stephen Lin.
\newblock Reppoints: Point set representation for object detection.
\newblock In {\em ICCV}, 2019.

\bibitem{zhang2020bridging}
Shifeng Zhang, Cheng Chi, Yongqiang Yao, Zhen Lei, and Stan~Z Li.
\newblock Bridging the gap between anchor-based and anchor-free detection via
  adaptive training sample selection.
\newblock In {\em CVPR}, 2020.

\bibitem{pang2019libra}
Jiangmiao Pang, Kai Chen, Jianping Shi, Huajun Feng, Wanli Ouyang, and Dahua
  Lin.
\newblock Libra r-cnn: Towards balanced learning for object detection.
\newblock In {\em CVPR}, 2019.

\bibitem{zhang2020dynamic}
Hongkai Zhang, Hong Chang, Bingpeng Ma, Naiyan Wang, and Xilin Chen.
\newblock Dynamic r-cnn: Towards high quality object detection via dynamic
  training.
\newblock In {\em ECCV}, 2020.

\bibitem{shao2018crowdhuman}
Shuai Shao, Zijian Zhao, Boxun Li, Tete Xiao, Gang Yu, Xiangyu Zhang, and Jian
  Sun.
\newblock Crowdhuman: A benchmark for detecting human in a crowd.
\newblock {\em arXiv preprint arXiv:1805.00123}, 2018.

\end{thebibliography}

\end{document}